\documentclass[11pt,a4paper]{article}
\usepackage[utf8]{inputenc}
\PassOptionsToPackage{breaklinks}{hyperref}
\usepackage[hyperref]{acl2020}
\usepackage{times}
\usepackage{latexsym}
\usepackage[capitalize,noabbrev]{cleveref}

\usepackage{booktabs}

\usepackage{soul}% \st , https://tex.stackexchange.com/questions/23711/strikethrough-text

\usepackage{microtype}

\aclfinalcopy % Uncomment this line for the final submission
\title{ELITR Non-Native Speech Translation at IWSLT 2020}
 
\def\normalspacer{\and}
\def\cuni{$^\dagger$}
\def\kit{$^\ddagger$}
\def\uedin{$^*$}
  \author{
Dominik Macháček\cuni\normalspacer
Jonáš Kratochvíl\cuni \normalspacer
Sangeet Sagar\cuni \normalspacer
\\
{\bf Mat{\' u}{\v s} {\v Z}ilinec\cuni \normalspacer
Ondřej Bojar\cuni\normalspacer
Thai-Son Nguyen\kit \normalspacer
Felix Schneider\kit \normalspacer
}%
\\
{\bf
Philip Williams\uedin \normalspacer
Yuekun Yao\uedin %\normalspacer
}
\\
\cuni{}Charles University,
\kit{}Karlsruhe Institute of Technology, \uedin{}University of Edinburgh
\\
\cuni{}\texttt{\{surname\}@ufal.mff.cuni.cz}, except \texttt{jkratochvil@ufal.mff.cuni.cz}, \\
\kit{}\texttt{\{firstname.lastname\}@kit.edu}, \\
\uedin{}\texttt{pwillia4@inf.ed.ac.uk,yyao2@exseed.ed.ac.uk}%, \texttt{rico.sennrich@ed.ac.uk}
}

\date{}

\begin{document}
\maketitle
\begin{abstract}
This paper is an ELITR system submission for the non-native speech translation task at IWSLT 2020. 
We describe systems for offline ASR, real-time ASR, and our cascaded approach to offline SLT and real-time SLT. 
We select our primary candidates from a pool of pre-existing systems, develop a new end-to-end general ASR system, and a hybrid ASR trained on non-native speech. The provided small validation set prevents us from carrying out a complex validation, but we submit all the unselected candidates for contrastive evaluation on the test set.
\end{abstract}

\section{Introduction}
%\XXX{DM: let's use "offline" and "online", not "off-line" or "on-line". I googled that any variant is correct, but we should be coherent.}

This paper describes the submission of the EU project ELITR (European Live Translator)\footnote{\url{http://elitr.eu}} to the non-native speech translation task at IWSLT 2020 \citep{iwslt:2020}. 
It is a result of a collaboration of project partners Charles University (CUNI), Karlsruhe Institute of Technology (KIT), and University of Edinburgh (UEDIN), relying on the infrastructure provided to the project by PerVoice company.

The non-native speech translation shared task at IWSLT 2020 complements other IWSLT tasks by new challenges. Source speech is non-native English. It is spontaneous, sometimes disfluent, and some of the recordings come from a particularly noisy environment. The speakers often have a significant non-native accent. In-domain training data are not available. They consist only of native out-domain speech and non-spoken parallel corpora. The validation data are limited to 6 manually transcribed documents, from which only 4 have reference translations. The target languages are Czech and German.

The task objectives are quality and simultaneity, unlike the previous tasks, which focused only on the quality. Despite the complexity, the resulting systems can be potentially appreciated by many users attending an event in a language they do not speak or having difficulties understanding due to unfamiliar non-native accents or unusual vocabulary.

We build on our experience from the past IWSLT and WMT tasks, see e.g. \citet{kit-iwslt19,Nguyen2017a,pham2017kit,WeteskoEtAl2019,BawdenEtAl2019,cuni2019wmt}. Each of the participating institutions has offered independent ASR and MT systems trained for various purposes and previous shared tasks.
We also create some new systems for this task and deployment for the purposes of the ELITR project.
%At our institutions we have many independent systems for ASR and machine translation from the task.
%\XXX{TODO -- the story of submission: It's just baseline, to see how far we can get with the tools we already have. Or, how did we try to improve it, if we really have something new.}
Our short-term motivation for this work is to connect the existing systems into a working cascade for SLT and evaluate it empirically, end-to-end. In the long-term, we want to advance state of the art in non-native speech translation.

%\XXX{To name systems consitently, always use the following macros, but remember to put empty curly brackets in the macro call, e.g. }\verb"\KIT{}".

%\XXX{Would it be possible to add a single-letter mark at the end of each system name to indicate if it was online or offline? Or are some systems used in both ways? DM: They're in both.}

\def\GoogleASR{Google}
\def\MicrosoftASR{Microsoft}
\def\KIT#1{KIT-h-large-lm#1} % \KIT1 \KIT2
\def\KITi{KIT-hybrid}
\def\KITStoS{KIT-seq2seq}
\def\KaldiMozilla{Kaldi-Mozilla}
\def\KaldiTed{Kaldi-TedLium}
\def\KaldiAMI{Kaldi-AMI}
\def\ThaiSonASR{KIT-seq2seq}
\def\PeterBase{Jasper-base}
\def\PeterFinetune{Jasper-finetune}
\def\PeterSlt{Jasper-from-slt}
\def\PeterBert{Jasper-bert}

% in the WER tables
\def\PeterAsrLm{Jasper-ASR-LM}
\def\PeterAsrSltLm{Jasper-ASR-SLT-LM}
\def\PeterAsr{\PeterBase}
\def\PeterBert{Jasper-Bert}
\def\PeterFromSlt{Jasper-ASR-SLT}
\def\PeterAsrWithLm{Jasper-ASR-LM-no-punct}
\def\PeterAsrSltWithLm{Jasper-ASR-SLT-LM-no-punct}

\def\yes{yes}
\def\no{no}

% \begin{table*}[t]
%     \centering
%     \small
%     \begin{tabular}{l|lllll}
%       \bf system &  \bf back-end &  \bf constrained &  \bf data &  \bf hours \\
%       \midrule
%       \KaldiTed{} & Kaldi &   \yes{}              & Ted-Lium     & 160\\
%       \KaldiMozilla{} & Kaldi & \yes{}             & Common Voice & 260\\
%       \KaldiAMI{} & Kaldi & \no{}                 & AMI          & 100\\
%       % \PeterSlt{} & NeMo & \no & Libri Speech \& Common Voice \& Czeng & 1600 \\
%       \GoogleASR{} & Google & \no{}                & -            & - \\
%       \MicrosoftASR{} & Microsoft & \no{}                & -            & - \\
% \end{tabular}
% \caption{Parameters of the considered ASR systems, \XXX{TODO: rather system, type hybrid/seq2seq, online/offline only, constrained, reference...}}
% \label{tab:asr-summary}
% \end{table*}

\section{Overview of Our Submissions}

This paper is a joint report for two primary submissions, for online and offline sub-track of the non-native simultaneous speech translation task. %They differ in the ASR component, but the rest of the work is joint for both sub-tracks.
% \XXX{The following must be fixed to match the primary submissions called CUNI-online and CUNI-offline. DM: What do you mean by CUNI-online and CUNI-offline? OB: Sorry, the names we are using for these submissions in the overview paper are ``ELITR'' and ``ELITR-offline'', }
% One was led by Dominik Macháček and focused on the SLT integration and end-to-end evaluation of the SLT cascade. The second primary submission was prepared primarily by Jonáš Kratochvíl and focused on Kaldi-based ASR solutions. Due to several common components, we decided to merge the reports into one paper.

% Dominik's team participated in all sub-tracks of the non-native simultaneous speech translation task. First, we collected all ASR systems that were available for us (\Cref{sec:asr-systems}) and evaluated them on the validation set. We selected the best candidate for offline ASR to serve as the source for offline SLT. Then, from the ASR systems, which are usable in online mode, we selected the best candidate for online ASR and as the basis for online SLT. 

% \XXX{DM: not true anymore. Rewrite.}
% Jonáš provides a second primary submission for offline ASR. It is hybrid ASR system in Kaldi (\cref{sec:kaldi-asr}). Although it currently performs worse on the validation set than the other primary system, it has a potential to achieve higher quality after some work and can be developed for use in online mode, unlike the first primary offline ASR.

First, we collected all ASR systems that were available for us (\Cref{sec:asr-systems}) and evaluated them on the validation set (\Cref{sec:asr-candidates}). We selected the best candidate for offline ASR to serve as the source for offline SLT. Then, from the ASR systems, which are usable in online mode, we selected the best candidate for online ASR and as a source for online SLT. 

In the next step (\Cref{sec:punct}), we punctuated and truecased the online ASR outputs of the validation set, segmented them to individual sentences, and translated them by all the MT systems we had available (\Cref{sec:mt}). We integrated the online ASRs and MTs into our platform for online SLT (\Cref{sec:slt,sec:mtwrapper}). We compared them using automatic MT quality measures and by simple human decision, to compensate for the very limited and thus unreliable validation set (\Cref{sec:mt-quality}). We selected the best candidate systems for each target language, for Czech and German.

Both best candidate MT systems are very fast (see \Cref{sec:mt-speed}).
%Their translation delay is approximately 300 milliseconds, in average (see \Cref{sec:mt-speed}). 
Therefore, we use them both for the online SLT, where the low translation time is critical, and for offline SLT.

In addition to the primary submissions, we included all the other candidate systems and some public services as contrastive submissions.

\section{Automatic Speech Recognition}
\label{sec:asr}

This section describes our automatic speech recognition systems and their selection.

\subsection{ASR Systems}
\label{sec:asr-systems}

% if we use table with ASR summary:
%See \Cref{tab:asr-summary} for a summary of ASR systems. 
%More details are in the following sections.

% if we don't:
We use three groups of ASR systems. They are described in the following sections.

%\XXX{TODO: the subsubsections should be revised. KIT online separately, KIT-seq2seq seperately, Kaldi-TedLium, Kaldi-AMI etc. should be distinct...}

\subsubsection{KIT ASR}
KIT has provided three hybrid HMM/ANN ASR systems and an end-to-end sequence-to-sequence ASR system.

The hybrid systems, called \KIT1{}, \KIT2{} and \KITi{}, were developed to run on the online low-latency condition, and differ in the use of the language models.

The \KIT{} adopted a 4-gram language model which was trained on a large text corpus \citep{Nguyen2017a}, while the KIT-hybrid employed only the manual transcripts of the speech training data. We would refer the readers to the system paper by \citet{Nguyen2017a} for more information on the training data and the studies by \citet{nguyen2020low,niehues2018lowlatency} for more information about the online setup.

The end-to-end ASR, so-called KIT-seq2seq, followed the architecture and the optimizations described by \citet{nguyen2019improving}. It was trained on a large speech corpus, which is the combination of Switchboard, Fisher, LibriSpeech, TED-LIUM, and Mozilla Common Voice datasets. It was used solely without an external language model.

All KIT ASR systems are unconstrained because they use more training data than allowed for the task.

\def\onl{$^{*}$}

\def\Antrecorp{Antrecorp}
\def\antrecorp{\Antrecorp}
\def\AMIdomain{AMI}
\def\AMI{\AMIdomain}
\def\teddy{Teddy}
\def\autocentrum{Autocentrum}
\def\AMIa{AMIa}
\def\AMIb{AMIb}
\def\AMIc{AMIc}
\def\AMId{AMId}
\def\auditing{Auditing}
\def\Auditing{Auditing}
\begin{table*}[t]
    \centering
    \small
    \begin{tabular}{l|rrrr|r@{~}r|rrrrrrr}
domain    &    \multicolumn{4}{c|}{\AMI} &    \multicolumn{2}{c|}{\antrecorp}    &    \auditing    \\
document    &    \AMIa    &    \AMIb    &    \AMIc    &    \AMId    &    \teddy    &    \autocentrum    &    \auditing    \\
\midrule
%\PeterAsrLm    &    35.95    &    32.94    &    35.57    &    40.43    &    56.20    &    13.10    &    10.67    \\
%\PeterAsrSltLm    &    37.72    &    33.86    &    35.59    &    40.59    &    56.42    &    13.10    &    10.71    \\
%\PeterAsr    &    35.89    &    32.76    &    35.60    &    39.90    &    57.43    &    11.62    &    9.83    \\
%\PeterBert    &    36.69    &    33.82    &    36.50    &    39.63    &    56.76    &    12.87    &    9.60    \\
%\PeterFromSlt    &    36.73    &    33.22    &    35.70    &    39.69    &    56.87    &    10.93    &    10.22    \\
%\PeterAsrWithLm    &    37.58    &    33.66    &    35.32    &    40.60    &    56.65    &    14.01    &    11.00    \\
%\PeterAsrSltWithLm    &    37.71    &    33.83    &    35.67    &    40.45    &    56.31    &    12.87    &    10.71    \\
\KIT1    &    50.71    &    47.96    &    53.11    &    50.43    &    65.92    &    19.25    &    18.54    \\
\KIT2    &    47.82    &    41.71    &    42.10    &    45.77    &    75.87    &    28.59    &    19.81    \\
\KITi    &    40.72    &    38.45    &    41.09    &    43.28    &    58.99    &    21.04    &    21.44    \\
\ThaiSonASR    &    33.73    &    28.54    &    34.45    &    42.24    &    42.57    &    9.91    &    10.45    \\
\KaldiTed    &    42.44    &    38.56    &    41.83    &    44.36    &    61.12    &    18.68    &    22.81    \\
\KaldiMozilla    &    52.89    &    56.37    &    58.50    &    58.90    &    68.72    &    45.41    &    34.36    \\
\KaldiAMI    &    \st{28.01}    &    \st{23.04}    &    \st{26.87}    &    \st{29.34}    &    59.66    &    20.62    &    28.39    \\
\MicrosoftASR    &    53.72    &    52.62    &    56.67    &    58.58    &    87.82    &    39.64    &    24.22    \\
\GoogleASR    &    51.52    &    49.47    &    53.11    &    56.88    &    61.01    &    14.12    &    17.47    \\
%\midrule
%transcript words    &    1788    &    4868    &    3454    &    1614    &    171    &    174    &    528    \\
    \end{tabular}
    \caption{WER rates of individual documents in the development set. \KaldiAMI{} scores on \AMI{} domain are striked through because they are unreliable due to an overlap with the training data.}
    \label{tab:asr-doc-wer}
\end{table*}

\begin{table}[t]
    \centering
    \small
    \begin{tabular}{l@{~}l@{~}|@{~}r@{~}r@{~}r@{~}r}
   domain      & document & sents. & tokens & duration & references\\
   \midrule
  \Antrecorp   & \teddy{} & 11 & 171 & 1:15 & 2\\
  \Antrecorp   & \autocentrum & 12 & 174 & 1:06 & 2 \\
  \Auditing    & \auditing & 25 & 528 & 5:38 & 1 \\
  \AMI         & \AMIa & 220 & 1788 & 15:09 & 1 \\
  \AMI         & \AMIb & 614 & 4868 & 35:17 &0 \\
  \AMI         & \AMIc & 401 & 3454 & 24:06 &0\\
  \AMI         & \AMId & 281 & 1614 & 13:01 &0\\
    \end{tabular}
    \caption{The size of the development set \texttt{iwslt2020-nonnative-minidevset-v2}. The duration is in minutes and seconds. As ``references'' we mean the number of independent referential translations into Czech and German.}
    \label{tab:dev-set}
\end{table}

\subsubsection{Kaldi ASR Systems}
\label{sec:kaldi-asr}

% \XXX{R1: KIT-TEDLIUM: "adaptation techniques were finetuned" - which techniques? This
% sentence is not clear.
% }\XXX{
% TODO-ask Jonáš, Jonáš: I am not sure about this as I have not work with KIT systems for this challenge but I would guess language model + lexicon adaptation? Perhaps Sangeet is more competent to confirm this .
% DM: R1 means Kaldi-TEDLIUM, the part with "adaptation techniques...". I think R1 complains only on the formulation. It's not a question. Jonas: Ok, I have included citation where the finetuning techniques are described in more detail
% }\XXX{
% AMI: why was the overlap of the validation set not removed from the training
% data of the KALDI-AMI system?
% }\XXX{
% Done}

We used three systems trained in the Kaldi ASR toolkit \citep{Povey_ASRU2011}. These systems were trained on Mozilla Common Voice, TED-LIUM, and AMI datasets together with additional textual data for language modeling.  
\paragraph{\KaldiMozilla{}}
For \KaldiMozilla{}, we used the Mozilla Common Voice baseline Kaldi recipe.\footnote{\url{https://github.com/kaldi-asr/kaldi/tree/master/egs/commonvoice/s5}} The training data consist of 260 hours of audio. The number of unique words in the lexicon is 7996, and the number of sentences used for the baseline language model is 6994, i.e., the corpus is very repetitive. We first train the GMM-HMM part of the model, where the final number of hidden states for the HMM is 2500, and the number of GMM components is 15000. We then train the chain model, which uses the Time delay neural network (TDNN) architecture \citep{peddinti2015time} together with the Batch normalization regularization and ReLU activation. We use MFCC features to represent audio frames, and we concatenate them with the 100-dimensional I-vector features for the neural network training. We recompile the final chain model with CMU lexicon to increase the model capacity to 127384 words and 4-gram language model trained with SRILM \citep{stolcke2002srilm} on 18M sentences taken from English news articles.
%We use the SRILM \citep{stolcke2002srilm} language modeling toolkit to train this 4-gram model. 

\paragraph{\KaldiTed{}} serves as another baseline, trained on 130 hours of TED-LIUM data \citep{rousseau2012ted} collected before the year 2012. The \KaldiTed{} model was developed by the University of Edinburgh and was fully described by \citet{DBLP:journals/corr/abs-1906-11521}. This model was primarily developed for discriminative acoustic adaptation to domains distinct from the original training domain. It is achieved by reusing the decoded lattices from the first decoding pass and by finetuning for TED-LIUM development and test set.
%described in \citet{DBLP:journals/corr/abs-1906-11521} in more detail. The modwere finetuned on the TED-LIUM development and test set. 
The setup follows the 
%\XXX{OB: check if the ``1f'' is not a typo. What does it mean? Jonas: it is not a typo, in Kaldi there are multiple available setups for TedLium (different NN architectures,...) and 1f is one of them...}
Kaldi 1f TED-LIUM recipe. The architecture is similar to \KaldiMozilla{} and uses a combination of TDNN layers with batch normalization and ReLU activation. The input features are MFCC and I-vectors. 

\paragraph{\KaldiAMI{}}
was trained on the 100 hours of the AMI data, which comprise of staged meeting recordings \citep{ami-corpus}. These data were recorded mostly by non-native English speakers with a different microphone and acoustic environment conditions. The model setup used follows the Kaldi 1i ami recipe. \KaldiAMI{} cannot be reliably assessed on the AMI part of the development due to the overlap of training and development data. We have decided not to exclude this overlap so that we do not limit the amount of available training data for our model. 

%\subsubsection{\XXX{TODO Peter: Jasper ASR a.k.a \PeterSlt{}}}

\subsubsection{Public ASR Services} 
As part of our baseline models, we have used Google Cloud Speech-to-Text API\footnote{\url{https://cloud.google.com/speech-to-text}} and Microsoft Azure Speech to Text.\footnote{\url{https://azure.microsoft.com/en-us/services/cognitive-services/speech-to-text/}} Both of these services provide an API for transcription of audio files in WAV format, and they use neural network acoustic models. We kept the default settings of these systems. 

The Google Cloud system supports over 100 languages and several types of English dialects (such as Canada, Ireland, Ghana, or the United Kingdom). For decoding of the development and test set, we have used the United Kingdom English dialect option. The system can be run either in real-time or offline mode. We have used the offline option for this experiment. 

The Microsoft Azure Bing Speech API supports fewer languages than Google Cloud ASR but adds more customization options of the final model. It can be also run both in real-time or offline mode. For the evaluation, we have used the offline mode and the United Kingdom English (en-GB) dialect. 

%\XXX{Which mode was used? DM: I hope it's correct. Jonáš knows. Jonas: Yes it was the offline option}
%We use the latter.

\subsection{Selection of ASR Candidates}
\label{sec:asr-candidates}

\begin{table}[t]
    \centering
    \small
    \begin{tabular}{l|r@{~}r@{~}r|r}
 & \multicolumn{3}{c|}{WER weighted average}    & avg                            \\
    &    \AMI    &    \antrecorp    &    \auditing    &    domain \\
\midrule
%\PeterAsrLm    &    35.20    &    34.46    &    10.67    &    26.78    \\
%\PeterAsrSltLm    &    35.88    &    34.57    &    10.71    &    27.06    \\
%\bf \PeterAsr$^2$    &    35.06    &    34.33    &    9.83    &    26.40    \\
%\PeterBert    &    35.85    &    34.62    &    \bf9.60    &    26.69    \\
%\PeterFromSlt    &    35.38    &    33.70    &    10.22    &    26.43    \\
%\PeterAsrWithLm    &    35.70    &    35.14    &    11.00    &    27.28    \\
%\PeterAsrSltWithLm    &    35.88    &    34.40    &    10.71    &    27.00    \\
\bf \ThaiSonASR$^1$    &    \bf 32.96    &    \bf 26.10    &    \bf 10.45    &    \bf23.17    \\
\KaldiTed    &    40.91    &    39.72    &    22.81    &    34.48    \\
\KaldiMozilla    &    56.82    &    56.96    &    34.36    &    49.38    \\
\KaldiAMI    &    \st{25.79}    &    39.97    &    28.39    &    31.38    \\
\MicrosoftASR    &    54.80    &    63.52    &    24.22    &    47.51    \\
\GoogleASR    &    51.88    &    37.36    &    17.47    &    35.57    \\
\midrule
\bf \KIT1    &    50.24    &    42.38    &    \bf 18.54\rlap{$^2$}    &    37.05    \\
\KIT2    &    43.32    &    52.02    &    19.81    &    38.38    \\
\bf \KITi    &    \bf 40.24\rlap{$^1$}    &    \bf 39.85\rlap{$^1$}    &    21.44    &    \bf 33.84    \\
    \end{tabular}
    \caption{Weighted average WER for the domains in validation set, and their average. The top line-separated group are offline ASR systems, the bottom are online. Bold numbers are the lowest considerable WER in the group. \KaldiAMI{} score on \AMI{} is not considered due to overlap with training data. Bold names are the primary (marked with $^1$) and secondary (marked with $^2$) candidates.}
    \label{tab:asr-wavg}
\end{table}

%\XXX{DM: the AMI-AMI should be striked out in the table. This section needs reformulation...}
% OB: no need to reformulate

We processed the validation set with all the ASR systems, evaluated WER, and summarized them in \Cref{tab:asr-doc-wer}. The validation set (\Cref{tab:dev-set}) contains three different domains with various document sizes, and the distribution does not fully correspond to the test set. The \AMI{} domain is not present in the test set at all, but it is a part of \KaldiAMI{} training data. Therefore, a simple selection by an average WER on the whole validation set could favor the systems which perform well on the \AMI{} domain, but they could not be good candidates for the other domains. 

In \Cref{tab:asr-wavg}, we present the weighted average of WER in the validation domains. We weight it by the number of gold transcription words in each of the documents. We observe that \KaldiAMI{} has a good performance on the \AMI{} domain, but it is worse on the others. We assume it is overfitted for this domain, and therefore we do not use it as the primary system.

For offline ASR, we use \ThaiSonASR{} as the primary system because it showed the lowest error rate on the averaged domain.
%\XXX{Now the question is if we are shifting \PeterAsr{} to the other paper...}
%\PeterAsr{} has the second lowest score, and the lowest score on \auditing{}, so we use it as a secondary candidate, and we use it as a source for offline SLT MT candidates.

The online ASR systems can exhibit somewhat lower performance than offline systems.
%generally \XXX{OB: what exactly do you mean by that? Within this task? Or really generally? DM: really generally. anyway, it shouldn't be here...} perform much worse than the offline end-to-end ASR, but they are substantially faster for use in online mode.
We select \KIT1 as the primary online ASR candidate for \auditing{}, and \KITi{} as primary for the other domains.

Our second primary offline ASR is \KaldiAMI{}. 
%\XXX{DM: Jonáši, ověřit, a odůvodnit, kterej z těch kaldi modelů se vybere.}

% \subsection{Online ASR}
% \label{sec:online-asr}

% \XXX{online are only \KIT1, \KIT2 and \KITi{}. Based on that, we selected...}

% \XXX{The baseline ASR, en-EU-lecture\_KIT-hybrid, is \citet{Nguyen2017a}.}
% \XXX{TODO Thai-Son: description or reference of en-EU-iwslt20\_1 and 2}

\section{Punctuation and Segmentation}
\label{sec:punct}

%\XXX{Important note for co-authors:
%Punctuation is adding .!?" and capitalization, it's by Felix or Sangeet's "segmentation" workers. We use the term incorrectly. Segmentation is splitting to sentences (or any segments) for MT, by online-text-flow-events or mt-wrapper, newly by MosesSentenceSegmenter. KIT MTs have different segmentation, I don't know how you do it.}

All our ASR systems output unpunctuated, often all lowercased text. The MT systems are designed mostly for individual sentences with proper casing and punctuation. To overcome this, we first insert punctuation and casing to the ASR output. Then, we split it into individual sentences by the punctuation marks by a rule-based language-dependent Moses sentence splitter \citep{moses}.

Depending on the ASR system, we use one of two possible punctuators. Both of them are usable in online mode.
% KIT only in online mode, but it's not important...

\subsection{KIT Punctuator}

% \XXX{R1:
% Punctuator systems: can you maybe explain how these seq2seq systems deal with
% long input? Are there any restrictions, do you rely on acoustic segmentation?
% What if someone talks long without pauses?
% TODO-Felix: It's not a seq2seq system, that's how :) I added some explanation of that.
% }

The KIT ASR systems use an NMT-based model to insert punctuation and capitalization in an otherwise unsegmented lowercase input stream \cite{cho2012segmentation, cho2015punctuation}. The system is a monolingual translation system that translates from raw ASR output to well-formed text by converting words to upper case, inserting punctuation marks, and dropping words that belong to disfluency phenomena. It does not use the typical sequence-to-sequence approach of machine translation. However, it considers a sliding window of recent (uncased) words and classifying each one according to the punctuation that should be inserted and whether the word should be dropped for being a part of disfluency. This gives the system a constant input and output size, removing the need for a sequence-to-sequence model.

While inserting punctuation is strictly necessary for MT to function at all, inserting capitalization and removing disfluencies improves MT performance by making the test case more similar to the MT training conditions \cite{cho2017nmt}.

\subsection{BiRNN Punctuator}

%\XXX{
%R1: BiRNN punctuator: why is it trained on the CzEng corpus (English side)? This is
%a monolingual task, you could have selected something closer to the spoken
%language domain easily. TODO-Sangeet
%}

For other systems, we use a bidirectional recurrent neural network with an attention-based mechanism by \citet{tilk2016} to restore punctuation in the raw stream of ASR output. The model was trained on 4M English sentences from CzEng 1.6 \citep{czeng16:2016} data and a vocabulary of 100K most frequently occurring words.
We use CzEng because it is a mixture of domains, both originally spoken, which is close to the target domain, and written, which has richer vocabulary, and both original English texts and translations, which we also expect in the target domain.
The punctuated transcript is then capitalized using an English tri-gram truecaser by \citet{Lita2003}. The truecaser was trained on 2M English sentences from CzEng.
%\XXX{DM: which truecasing model do we use, Sangeet? n-gram? for which n?- uni , bi and tri all}
%\XXX{I can write abaout it. It uses all by the way}
% thanks, it's solved
% the trucaser was trained using Greed strategy. The greedy strategy uses casing of uni, bi, tri- grams from given training data to train the dictionary and replacing the word token with the highest probable casing. This resulted in a small model size making the overall punctuation restoration task faster.
% you can use it if required.
% the word tri-gram is enough.
% Thank You. Its resolved then.
% yes. just remember, that the word is truecaser, with e, not trucaser. Yes yes, I will use your suggestion and fix in Peter's Paper.

\section{Machine Translation}

This section describes the translation part of SLT.

\subsection{MT Systems}
\label{sec:mt}

\def\to{$\rightarrow$}
\def\doubleto{$\leftrightarrow$}
\def\csCZ{WMT19 Marian}
\def\csIWSLT{IWSLT19}
\def\csLindat{WMT18 T2T}
\def\rb{OPUS-A}
\def\rbnew{OPUS-B}
\def\rbMatus{T2T-multi}
\def\rbMatusBig{T2T-multi-big}
\def\deDE{de-LSTM}
\def\deLindat{de-T2T}

\begin{table*}[]
    \centering
    \small
    \begin{tabular}{l|llllll}
 \bf   system &  \bf back-end &  \bf source-target &  \bf constrained &  \bf reference \\
 \midrule
\csCZ & Marian & en\to{}cs & \no{} & \citet{cuni2019wmt}, \Cref{sec:wmt19} \\
\csLindat{} & T2T & en\to{}cs & \no{} & \citet{cuni2019wmt}, \Cref{sec:wmt19} \\
\csIWSLT{} & Marian & en\to{}cs & \no{} & \citet{WeteskoEtAl2019}, \Cref{sec:iwslt19} \\
\rb {} & Marian & en\doubleto $\{$cs,de+5 l.$\}$ & \no{}  & \Cref{sec:marianmulti} \\
\rbnew {} & Marian & en\doubleto $\{$cs,de+39 l.$\}$ & \no{}  & \Cref{sec:marianmulti} \\
\rbMatus & T2T & en\doubleto $\{$cs,de,en+39 l.$\}$ & \no{} & \Cref{sec:matusmodels} \\
\rbMatusBig{} & T2T & en\doubleto $\{$cs,de,en+39 l.$\}$ & \no{} & \Cref{sec:matusmodels} \\
\deDE & NMTGMinor & en\to{}de & no & \citet{dessloch-etal-2018-kit}, \Cref{sec:mt-deDE} \\
\deLindat & T2T & en\to{}de & \yes & \Cref{sec:deLindat} \\
    \end{tabular}
    \caption{The summary of our MT systems.}
    \label{tab:mt-systems}
\end{table*}

See \Cref{tab:mt-systems} for the summary of the MT systems. All except \deDE{} are Transformer-based neural models using Marian \citep{mariannmt} or Tensor2Tensor \citep{t2t} back-end. All of them, except \deLindat{}, are unconstrained because they are trained not only on the data sets allowed in the task description, but all the used data are publicly available.

%Four of the systems (\rb, \rbnew, \rbMatus, and \rbMatusBig) are trained on multi-lingual English-to-something

\subsubsection{WMT Models}
\label{sec:wmt19}

\csCZ{} and \csLindat{} models are Marian and T2T single-sentence models from \citet{cuni2019wmt} and \citet{popel-2018-cuni}. \csLindat{} was originally trained for the English-Czech WMT18 news translation task, and reused in WMT19. \csCZ{} is its reimplementation in Marian for WMT19. The T2T model has a slightly higher quality on the news text domain than the Marian model. The Marian model translates faster, as we show in \Cref{sec:mt-speed}.

\subsubsection{\csIWSLT{} Model}
\label{sec:iwslt19}

The \csIWSLT{} system is an ensemble of two English-to-Czech Transformer Big models trained using the Marian toolkit.
The models were originally trained on WMT19 data and then finetuned on MuST-C TED data.
The ensemble was a component of Edinburgh and Samsung's submission to the IWSLT19 Text Translation task.
See Section 4 of \citet{WeteskoEtAl2019} for further details of the system.

\subsubsection{OPUS Multi-Lingual Models}
\label{sec:marianmulti}

The OPUS multilingual systems are one-to-many systems developed within the ELITR project.
Both were trained on data randomly sampled from the OPUS collection~\cite{TIEDEMANN12.463}, although they use distinct datasets.
\rb{} is a Transformer Base model trained on 1M sentence pairs each for 7 European target languages: Czech, Dutch, French, German, Hungarian, Polish, and Romanian.
\rbnew{} is a Transformer Big model trained on a total of 231M sentence pairs covering 41 target languages that are of particular interest to the project\footnote{The 41 target languages include all EU languages (other than English) and 18  languages that are official languages of EUROSAI member countries. Specifically, these are Albanian, Arabic, Armenian, Azerbaijani, Belorussian, Bosnian, Georgian, Hebrew, Icelandic, Kazakh, Luxembourgish, Macedonian, Montenegrin, Norwegian, Russian, Serbian, Turkish, and Ukrainian.}
After initial training, \rbnew{} was finetuned on an augmented version of the dataset that includes partial sentence pairs, artificially generated by truncating the original sentence pairs (similar to \citealp{niehues2018lowlatency}).
We produce up to 10 truncated sentence pairs for every one original pair.

\subsubsection{T2T Multi-Lingual Models}
\label{sec:matusmodels}

%\XXX{Reviewer 2: "The models were trained in a one-to-many,
%many-to-one fashion". I assume many-to-one is a back-translation approach, but
%it isn't clear to me. Please cite or elaborate. TODO Matúš. Explain that you mean one source into many target languages as in Johnson et al. -- OK?}

\rbMatus{} and \rbMatusBig{} are respectively Transformer and Transformer Big models trained on a Cloud TPU based on the default T2T hyperparameters, with the addition of target language tokens as in \citet{johnson-etal-2017-googles}. The models were trained with a shared vocabulary on a dataset of English-to-many and many-to-English sentence pairs from \rbnew{} containing 42 languages in total, making them suitable for pivoting. The models do not use finetuning.

\subsubsection{\deLindat{}}
\label{sec:deLindat}

\deLindat{} translation model is based on a Tensor2Tensor translation model model using training hyper-parameters similar to \citet{popel2018trainingtips}.
The model is trained using all the parallel corpora provided for the English-German WMT19 News Translation Task, without back-translation.
We use the last training checkpoint during model inference.
To reduce the decoding time, we apply greedy decoding instead of a beam search.

\subsubsection{KIT Model}
\label{sec:mt-deDE}

KIT's translation model is based on an LSTM encoder-decoder framework with attention \cite{pham2017kit}. As it is developed for our lecture translation framework \cite{muller2016lecture}, it is finetuned for lecture content. In order to optimize for a low-latency translation task, the model is also trained on partial sentences in order to provide more stable translations \cite{niehues2016dynamic}.

\subsection{ELITR SLT Platform}
\label{sec:slt}

We use a server called Mediator for the integration of independent ASR and MT systems into a cascade for online SLT. It is a part of the ELITR platform for simultaneous multilingual speech translation \citep{ELITR-ELG}. The workers, which can generally be any audio-to-text or text-to-text processors, such as ASR and MT systems, run inside of their specific software and hardware environments located physically in their home labs around Europe. They connect to Mediator and offer a service. A client, often located in another lab, requests Mediator for a cascade of services, and Mediator connects them. This platform simplifies the cross-institutional collaboration when one institution offers ASR, the other MT, and the third tests them as a client. The platform enables using the SLT pipeline easily in real-time.

\subsection{MT Wrapper}
\label{sec:mtwrapper}

%\XXX{@Dominik: please move the source of \Cref{tab:mt-cs-bleu} here, hopefully the table will move some pages ahead.}

% the tables are edited here https://docs.google.com/spreadsheets/d/1LmRkCVPesFlmM9OFqcnqdDyNpUImKZL_bvZYo5ZXXAM/edit?usp=sharing
% in the sheet paper-latex-tables
% it's better to edit it there, copypaste here and then add \midrule
\begin{table*}[t]
    \centering
    \footnotesize
    \begin{tabular}{l|l|r|r@{~}r@{~}r|r}
MT    & document        &    gold &    \KITi{}    &    \KIT1    &    \KIT2    &    avg KIT \\
\midrule
\rbnew    &    \teddy    &    42.846    &    2.418    &    2.697    &    1.360    &    2.158    \\
\csIWSLT    &    \teddy    &    51.397    &    1.379    &    2.451    &    1.679    &    1.836    \\
\csCZ    &    \teddy    &    49.328    &    1.831    &    1.271    &    1.649    &    1.584    \\
\csLindat    &    \teddy    &    54.778    &    1.881    &    1.197    &    1.051    &    1.376    \\
\rb    &    \teddy    &    25.197    &    1.394    &    1.117    &    1.070    &    1.194    \\
\rbMatus    &    \teddy    &    36.759    &    1.775    &    0.876    &    0.561    &    1.071    \\
\csLindat    &    \autocentrum    &    42.520    &    12.134    &    13.220    &    14.249    &    13.201    \\
\csCZ    &    \autocentrum    &    39.885    &    10.899    &    10.695    &    12.475    &    11.356    \\
\rbnew    &    \autocentrum    &    29.690    &    12.050    &    10.873    &    9.818    &    10.914    \\
\csIWSLT    &    \autocentrum    &    37.217    &    9.901    &    8.996    &    8.900    &    9.266    \\
\rb    &    \autocentrum    &    30.552    &    9.201    &    9.277    &    8.483    &    8.987    \\
\rbMatus    &    \autocentrum    &    20.011    &    6.221    &    2.701    &    3.812    &    4.245    \\
\midrule                                                    
\csIWSLT    &    \AMIa    &     \bf 22.878    &    5.377    &    2.531    &     3.480    &     \bf 3.796    \\
\csLindat    &    \AMIa    &    21.091    &    5.487    &    2.286    &    3.411    &    3.728    \\
\csCZ    &    \AMIa    &    22.036    &    4.646    &    2.780    &    3.739    &    3.722    \\
\rbnew    &    \AMIa    &    19.224    &    4.382    &    3.424    &    2.672    &    3.493    \\
\rb    &    \AMIa    &    15.432    &    3.131    &    2.431    &    2.500    &    2.687    \\
\rbMatus    &    \AMIa    &    13.340    &    2.546    &    2.061    &    1.847    &    2.151    \\
\midrule                                                    
\csIWSLT    &    \auditing    &     \bf 9.231    &    1.096    &    3.861    &    2.656    &     \bf 2.538    \\
\rbnew    &    \auditing    &    6.449    &    1.282    &    3.607    &    2.274    &    2.388    \\
\rb    &    \auditing    &    8.032    &    1.930    &    4.079    &    0.900    &    2.303    \\
\csCZ    &    \auditing    &    8.537    &    1.087    &    3.571    &    1.417    &    2.025    \\
\csLindat    &    \auditing    &    9.033    &    1.201    &    2.935    &    1.576    &    1.904    \\
\rbMatus    &    \auditing    &    3.923    &    1.039    &    1.318    &    1.110    &    1.156    \\
    \end{tabular}
    \caption{Validation BLEU scores in percents (range 0-100) for SLT into Czech from ASR sources. The column ``gold'' is translation from the gold transript. It shows the differences between MT systems, but was not used in validation.
    %\rbMatusBig{} is missing because it was too slow for online ASR.
    }
    \label{tab:mt-cs-bleu}
\end{table*}

The simultaneous ASR incrementally produces the recognition hypotheses and gradually improves them. The machine translation system translates one batch of segments from the ASR output at a time. If the translation is not instant, then some ASR hypotheses may be outdated during the translation and can be skipped. We use a program called MT Wrapper for connecting the output of self-updating ASR with non-instant NMT systems.

MT Wrapper has two threads. The receiving thread segments the input for our MTs into individual sentences, saves the input into a buffer, and continuously updates it. The translating thread is a loop that retrieves the new content from the buffer. If a segment has been translated earlier in the current process, it is outputted immediately. Otherwise, the new segments are sent in one batch to the NMT system, stored to a cache and outputted.

For reproducibility, the translation cache is empty at the beginning of a process, but in theory it could be populated by a translation memory. The cache significantly reduces the latency because the punctuator often oscillates between two variants of casing or punctuation marks within a short time.

MT Wrapper has a parameter to control the stability and latency. It can mask the last $k$ words of incomplete sentences from the ASR output, as in \citet{ma-etal-2019-stacl} and \citet{Arivazhagan2019ReTranslationSF}, considering only the currently completed sentences, or only the ``stable'' sentences, which are beyond the ASR and punctuator processing window and never change.
We do not tune these parameters in the validation. We do not mask any words or segments in our primary submission, but we submit multiple non-primary systems differing in these parameters.

\subsection{Quality Validation}
\label{sec:mt-quality}

For comparing the MT candidates for SLT, we processed the validation set by three online ASR systems, translated them by the candidates, aligned them with reference by mwerSegmenter \citep{segmentation:matusov:2005:iwslt} and evaluated the BLEU score \citep{post-2018-call,papineni-etal-2002-bleu} of the individual documents. However, we were aware that the size of the validation set is extremely limited (see \Cref{tab:dev-set}) and that the automatic metrics as the BLEU score estimate the human judgment of the MT quality reliably only if there is a sufficient number of sentences or references. It is not the case of this validation set.

Therefore, we examined them by a simple comparison with source and reference. We realized that the high BLEU score in the \autocentrum{} document is induced by the fact that one of the translated sentences matches exactly matches a reference because it is a single word ``thanks''. This sentence increases the average score of the whole document, although the rest is unusable due to mistranslated words. The ASR quality of the two \Antrecorp{} documents is very low, and the documents are short. Therefore we decided to omit them in comparison of the MT candidates.

We examined the differences between the candidate translations on the \auditing{} document, and we have not seen significant differences, because this document is very short. The \AMIa{} document is longer, but it contains long pauses and many isolated single-word sentences, which are challenging for ASR. The part with a coherent speech is very short.

Finally, we selected the MT candidate, which showed the highest average BLEU score on the three KIT online ASR systems both on \auditing{} and \AMIa{} document because we believe that averaging the three ASR sources shows robustness against ASR imperfections. See \Cref{tab:mt-cs-bleu} and \Cref{tab:mt-de-bleu} for the BLEU scores on Czech and German. The selected candidates are \csIWSLT{} for Czech and \rbnew{} for German. However, we also submit all other candidates as non-primary systems to test them on a significantly larger test set.
We use these candidates both for online and offline SLT. %because at the time of submission, the final offline ASR candidate was not available.

\begin{table*}[t]
    \centering
    \footnotesize
    \begin{tabular}{l|l|r|r@{~}r@{~}r|r}
MT    &    document    &    gold    &    \KITi    &    \KIT1    &    \KIT2    &    avg    KIT \\
\midrule                                                    
\deLindat{}    &    \teddy    &    45.578    &    2.847    &    3.181    &    1.411    &    2.480    \\
\rb    &    \teddy    &    29.868    &    1.873    &    1.664    &    1.139    &    1.559    \\
\deDE{}    &    \teddy    &    3.133    &    2.368    &    2.089    &    1.254    &    1.904    \\
\rbnew{}    &    \teddy    &    41.547    &    2.352    &    1.878    &    1.454    &    1.895    \\
\rbMatus{}    &    \teddy    &    31.939    &    1.792    &    3.497    &    1.661    &    2.317    \\
\deLindat{}    &    \autocentrum    &    36.564    &    9.031    &    6.229    &    3.167    &    6.142    \\
\rb    &    \autocentrum    &    26.647    &    8.898    &    13.004    &    2.324    &    8.075    \\
\deDE{}    &    \autocentrum    &    19.573    &    10.395    &    13.026    &    2.322    &    8.581    \\
\rbnew{}    &    \autocentrum    &    28.841    &    10.153    &    12.134    &    9.060    &    10.449    \\
\rbMatus{}    &    \autocentrum    &    22.631    &    8.327    &    8.708    &    6.651    &    7.895    \\
\midrule                                                    
\deLindat{}    &    \AMIa    &    34.958    &    8.048    &    5.654    &    7.467    &    7.056    \\
\rb    &    \AMIa    &    30.203    &    7.653    &    5.705    &    5.899    &    6.419    \\
\deDE{}    &    \AMIa    &    31.762    &    7.635    &    6.642    &    1.843    &    5.373    \\
\rbnew{}    &    \AMIa    &    38.315    &    8.960    &    7.613    &    6.837    &    7.803    \\
\rbMatus{}    &    \AMIa    &    28.279    &    6.202    &    3.382    &    3.869    &    4.484    \\
\midrule                                                    
\deLindat{}    &    \auditing    &    38.973    &    11.589    &    17.377    &    18.841    &    15.936    \\
\rb    &    \auditing    &    38.866    &    10.355    &    19.414    &    18.540    &    16.103    \\
\deDE{}    &    \auditing    &    21.780    &    10.590    &    12.633    &    11.098    &    11.440    \\
\rbnew{}    &    \auditing    &    38.173    &    10.523    &    18.237    &    17.644    &    15.468    \\
\rbMatus{}    &    \auditing    &    22.442    &    7.896    &    8.664    &    11.269    &    9.276    \\
    \end{tabular}
    \caption{Validation BLEU scores in percents (range 0-100) for MT translations into German from ASR outputs and from the gold transcript.}
    \label{tab:mt-de-bleu}
\end{table*}

\subsection{Translation Time}
\label{sec:mt-speed}

\def\plusm{&$\pm$&}
\begin{table}[]
    \centering
    \small
\begin{tabular}{l|r@{~}c@{}r}
\bf MT & \bf avg\plusm{}\bf std dev \\
\midrule
\rbMatus{} &    2876.52\plusm 1804.63 \\
\rbMatusBig{} &    5531.30\plusm 3256.81 \\
\bf \csIWSLT{} & 275.51\plusm 119.44 \\
\csCZ{} & 184.08\plusm 89.17 \\
\csLindat{} & 421.11\plusm201.64 \\
\bf \rbnew{} & 287.52\plusm 141.28 \\
\rb{} & 263.31\plusm 124.75 \\
\end{tabular}
    \caption{Average and standard deviation time for translating one batch in validation set, in milliseconds. Bold are the candidate systems for online SLT.}
    \label{tab:mt-speed}
\end{table}

We measured the average time, in which the MT systems process a batch of segments of the validation set (\Cref{tab:mt-speed}). If the ASR updates are distributed uniformly in time, than the average batch translation time is also the expected delay of machine translation. 
The shortest delay is almost zero; in cases when the translation is cached or for very short segments. The longest delay happens when an ASR update arrives while the machine is busy with processing the previous batch. The delay is time for translating two subsequent batches, waiting and translating.

We suppose that the translation time of our primary candidates is sufficient for real-time translation, as we verified in on online SLT test sessions.

%\XXX{DM: the rest of section could be removed:}
We observe differences between the MT systems. The size and the model type of \csCZ{} and \csLindat{} are the same (see \citealp{cuni2019wmt}), but they differ in implementation.

\csCZ{} is slightly faster than \csIWSLT{} model because the latter is an ensemble of two models. \rbnew{} is slower than \rb{} because the former is bigger. Both are slower than \csCZ{} due to multi-targeting and different preprocessing. \csCZ{} uses embedded SentencePiece \citep{kudo-richardson-2018-sentencepiece}, while the multi-target models use an external Python process for BPE \citep{Sennrich2016NeuralMT}. 
The timing may be affected also by different hardware.

At the validation time, \rbMatus{} and \rbMatusBig{} used suboptimal setup.

%\subsection{MT Candidates}

%\subsection{Non-Primary Systems}

%\section{Results}

%This section will be included after the evaluation on the test set, if accepted.\XXX{TDOO -- remove the whole section or insert it, if we receive the results}

%The end-to-end neural ASR of the other participating team, \citet{iwslt20-polak}, seems to have lower performance on the offline ASR than our primary \ThaiSonASR{} system on the validation set. 

\section{Conclusion}

We presented ELITR submission for non-native SLT at IWSLT 2020. We observe a significant qualitative difference between the end-to-end offline ASR methods and hybrid online methods. The component that constrains the offline SLT from real-time processing is the ASR, not the MT. 

We selected the best candidates from a pool of pre-existing and newly developed components, and submitted our primary submissions,
%(online ASR, offline ASR, online Czech SLT, offline Czech SLT, online German SLT, offline German SLT, offline Kaldi-AMI ASR\XXX{DM: really?}). 
although the size of the development set limits us from a reliable validation. Therefore, we submitted all our unselected candidates for contrastive evaluation on the test set. For the results, we refer to \citet{iwslt:2020}.

%\XXX{TODO -- depending how the story ends}

%\XXX{A very nasty way of saving space is to change authors lists to "First Person and others" which usually renders as et al.}
%\XXX{DM: It's not necessary, the length is up to 8 pages + unlimited references. We have 8 pages.}

\section*{Acknowledgments}

The research was partially supported by the grants
19-26934X (NEUREM3) of the Czech Science Foundation, % Ondrej GACR EXPRO
H2020-ICT-2018-2-825460 (ELITR) of the EU, % Sangeet and many others
398120 of the Grant Agency of Charles University,  % Dominik, Peter
and by SVV project number 260 575. % (SVV, Dominik)

%\scriptsize %% w don't need to save the space. It's 8 pages excluding unlimited references.
\bibliography{biblio}
\bibliographystyle{acl_natbib}

\end{document}